\documentclass[conference]{IEEEtran}
\IEEEoverridecommandlockouts

\usepackage{cite}
\usepackage{amsmath,amssymb,amsfonts}
\usepackage{algorithmic}
\usepackage{textcomp}
\usepackage{algorithm}
\usepackage{algorithmic}
\usepackage{float} 
\usepackage{xcolor}
\usepackage{newfloat}
\usepackage{listings}
\usepackage{multirow}
\usepackage{times}  
\usepackage{helvet}  
\usepackage{courier}  
\usepackage[hyphens]{url}  
\usepackage{graphicx} 
\def\BibTeX{{\rm B\kern-.05em{\sc i\kern-.025em b}\kern-.08em
    T\kern-.1667em\lower.7ex\hbox{E}\kern-.125emX}}
\begin{document}

\title{FLeNS: Federated Learning with Enhanced Nesterov-Newton Sketch
}

\author{\IEEEauthorblockN{Sunny Gupta}
\IEEEauthorblockA{\textit{Koita Centre for Digital Health} \\
\textit{Indian Institute of Technology, Bombay}\\
Mumbai, India \\
sunnygupta@iitb.ac.in}
\and
\IEEEauthorblockN{Mohit Jindal}
\IEEEauthorblockA{\textit{Department of Electrical Engineering} \\
\textit{Indian Institute of Technology, Bombay}\\
Mumbai, India \\
jindalmohit351@gmail.com}
\and
\IEEEauthorblockN{Pankhi Kashyap}
\IEEEauthorblockA{\textit{Koita Centre for Digital Health} \\
\textit{Indian Institute of Technology, Bombay}\\
Mumbai, India \\
pankhikashyap.research@gmail.com}
\and
\IEEEauthorblockN{Pranav Jeevan}
\IEEEauthorblockA{\textit{Department of Electrical Engineering} \\
\textit{Indian Institute of Technology, Bombay}\\
Mumbai, India \\
pjeevan@iitb.ac.in}
\and
\IEEEauthorblockN{Amit Sethi}
\IEEEauthorblockA{\textit{Department of Electrical Engineering} \\
\textit{Indian Institute of Technology, Bombay}\\
Mumbai, India \\
asethi@iitb.ac.in}

}

\maketitle
\begin{abstract}
Federated learning faces a critical challenge in balancing communication efficiency with rapid convergence, especially for second-order methods. While Newton-type algorithms achieve linear convergence in communication rounds, transmitting full Hessian matrices is often impractical due to quadratic complexity. We introduce Federated Learning with Enhanced Nesterov-Newton Sketch (FLeNS), a novel method that harnesses both the acceleration capabilities of Nesterov's method and the dimensionality reduction benefits of Hessian sketching. FLeNS approximates the centralized Newton's method without relying on the exact Hessian, significantly reducing communication overhead. By combining Nesterov's acceleration with adaptive Hessian sketching, FLeNS preserves crucial second-order information while preserving the rapid convergence characteristics. Our theoretical analysis, grounded in statistical learning, demonstrates that FLeNS achieves super-linear convergence rates in communication rounds - a notable advancement in federated optimization. We provide rigorous convergence guarantees and characterize tradeoffs between acceleration, sketch size, and convergence speed. Extensive empirical evaluation validates our theoretical findings, showcasing FLeNS's state-of-the-art performance with reduced communication requirements, particularly in privacy-sensitive and edge-computing scenarios. The code is available at https://github.com/sunnyinAI/FLeNS
\end{abstract}

\section{Introduction}

Federated Learning (FL) has emerged as an up-and-coming framework for large-scale machine learning tasks, by providing substantial benefits in terms of privacy protection and computational efficiency \cite{konecny2016federated, mcmahan2017communication, li2020federated, wei2021federated, li2023optimal}. By enabling decentralized model training without sharing raw data, FL addresses concerns of data sensitivity and ownership. However, a significant challenge in FL is balancing the tradeoff between fast convergence rates and minimizing communication overhead.

First-order optimization methods, such as FedAvg \cite{mcmahan2017communication} and FedProx \cite{li2020federated}, have been widely adopted in FL due to their efficiency in communicating only gradient information, thus preserving data privacy while handling heterogeneous data distributions. Despite their success, these methods exhibit slow convergence rates, typically achieving only sublinear convergence, \(O(1/t)\), where \(t\) represents communication rounds \cite{li2020convergence, karimireddy2020scaffold, pathak2020fedsplit}. Recent advances in convergence analysis \cite{li2020federated, karimireddy2020scaffold, pathak2020fedsplit, glasgow2022sharp} and generalization bounds \cite{mohri2019agnostic, li2023optimal, su2021non, yuan2022what} have solidified the theoretical foundations of these approaches. However, the inherent limitations of first-order methods in terms of slow convergence continue to pose significant challenges for complex, high-dimensional problems.

In contrast, second-order methods, such as Newton’s method, are known for their superior convergence rates in centralized settings, often achieving linear or super-linear convergence under mild conditions \cite{boyd2004convex, bottou2018optimization}. However, directly applying these methods in FL is impractical due to the high communication costs associated with transmitting local Hessian matrices. Various approximations, including BFGS \cite{broyden1970convergence}, L-BFGS \cite{liu1989limited}, inexact Newton \cite{dembo1982inexact}, Gauss-Newton \cite{schraudolph2002fast} and Newton sketch \cite{pilanci2017newton}, have been proposed to mitigate computational complexity, but adapting them to FL while maintaining communication efficiency remains a challenge.

In this paper, we propose \textbf{FLeNS} (Federated Learning with Enhanced Nesterov-Newton Sketch), a novel optimization algorithm that harnesses the acceleration capabilities of Nesterov's method \cite{nesterov1983method} and the dimensionality reduction benefits of Hessian sketching \cite{pilanci2017newton}. FLeNS efficiently approximates Newton’s method while significantly reducing communication complexity by leveraging adaptive Hessian sketching, which reduces the sketch size to the effective dimension of the Hessian matrix. This innovative approach addresses the key challenges of federated learning—ensuring fast convergence, reducing computational burden, and minimizing communication costs, all while maintaining robust generalization performance. FLeNS makes several key contributions to federated optimization:

    \textbf{On the algorithmic front:} FLeNS introduces a second-order federated optimization algorithm that integrates Nesterov’s acceleration with sketched Hessians to achieve superlinear convergence which effectively reduces the communication complexity, making it scalable in real-world federated environments.
    
     \textbf{On the statistical front:} FLeNS offers a rigorous theoretical framework that guarantees both convergence speed and communication efficiency. This balance enables the algorithm to scale well with large datasets without compromising accuracy ensuring that superlinear convergence can be maintained with minimal communication overhead.




\begin{figure*}[htbp]
    \centering
    \includegraphics[width=0.9\textwidth]{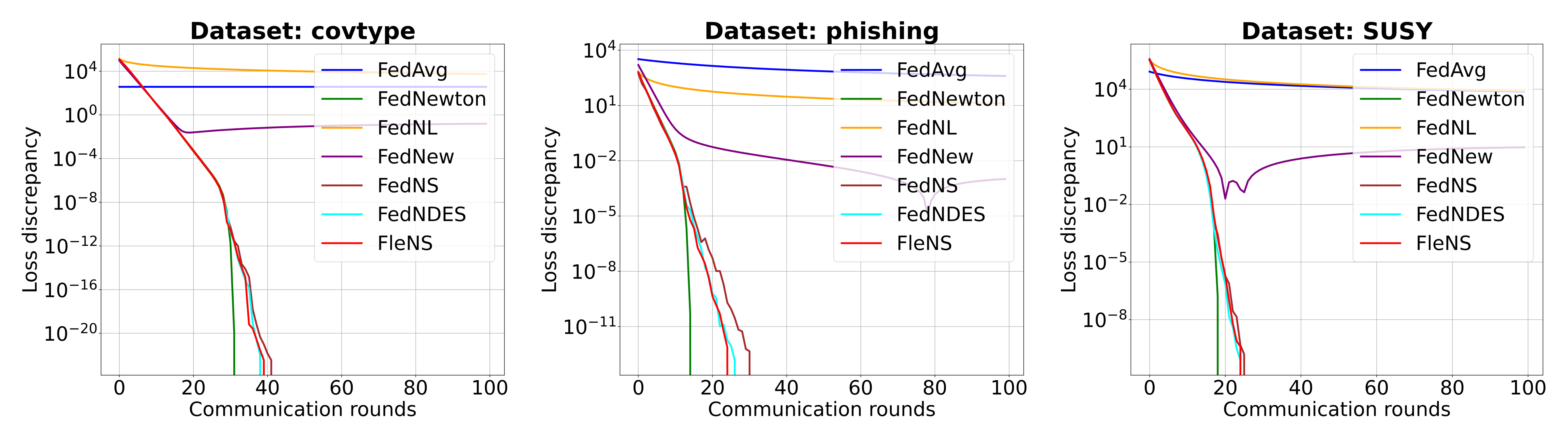}
    \caption{Loss discrepancy between the compared methods and the optimal learner as a function of the no. of communication rounds \( t \).}
    \label{fig1}
\end{figure*}
\begin{figure*}[htbp]
    \centering
    \includegraphics[width=0.9\textwidth]{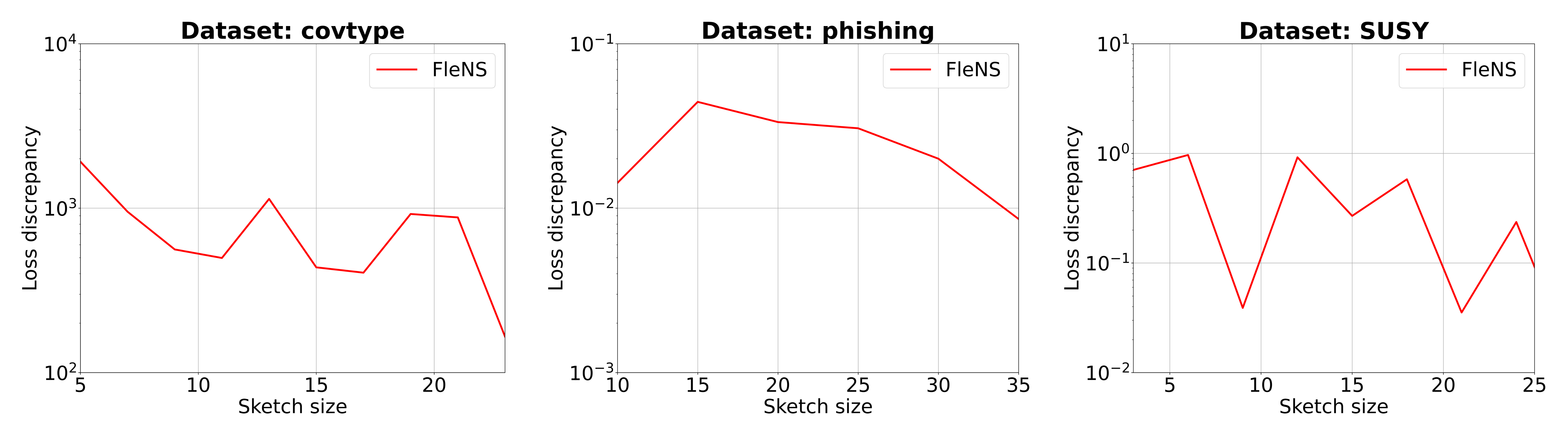}
    \caption{Loss discrepancy of the optimal learner as a function of the sketch size \( k \).}
    \label{fig2}
\end{figure*}
\section{Related Work}
In Federated Learning (FL), methods like FedAvg \cite{mcmahan2017communication} and FedProx \cite{li2020federated} are widely used due to their simplicity and communication efficiency. These algorithms only transmit local gradients of size \(M\), where \(M\) is the dimension of the feature space, resulting in a communication complexity of \(O(M)\). The convergence properties of FedAvg and FedProx have been thoroughly analyzed in \cite{li2020convergence}, with an iteration complexity of \(T = O(1/\delta)\) required to obtain a \(\delta\)-accurate solution. However, the sublinear convergence rates of these methods pose significant challenges for large-scale, high-dimensional problems, especially when dealing with non-i.i.d. data distributions across clients.

To accelerate convergence, \textbf{Newton Sketch} \cite{pilanci2017newton} was introduced as a second-order method that approximates Newton’s method using sketching techniques to reduce computational and communication burdens. Newton Sketching has been further optimized by methods such as Newton-LESS \cite{derezinski2021newton, lacotte2020effective, lacotte2021adaptive} reducing the effective dimension of the Hessian matrix, reducing communication complexity from \(O(M^2)\) to \(O(kM)\), where \(k \ll M\) is the sketch size.

Several \textbf{Newton-type federated algorithms} have been proposed to overcome the slow convergence of gradient-based methods in FL. Distributed Newton \cite{ghosh2020distributed} and Local Newton \cite{gupta2021localnewton} replace SGD with Newton’s method on local machines, accelerating convergence for local models. FedNew \cite{elgabli2022fednew} introduced a one-pass ADMM to calculate local directions, while an approximate Newton method is used to update the global model. FedNL \cite{safaryan2022fednl} sends compressed local Hessian updates to the global server to perform a global Newton step, reducing the communication overhead. Additionally, SHED \cite{dal2024shed} uses eigendecomposition of local Hessian matrices, incrementally updating eigenvector-eigenvalue pairs to the global server to recover the Hessian for performing Newton’s method globally.

In this paper, we propose \textbf{FLeNS}, which builds on these developments by extending Newton Sketch to FL. Our methods achieve super-linear convergence with a communication complexity of \(O(kM)\), where \(k\) is the effective dimension of the Hessian. By leveraging techniques like Nesterov-Newton Sketch, we minimize communication rounds while maintaining the fast convergence properties of Newton’s methods.
\section{Background}
A standard federated learning (FL) system consists of a global server and \( m \) local nodes, each with its dataset \( D_j = \{(x_{ij}, y_{ij})\}_{i=1}^{n_j} \), where \( \forall j \in [m] \), the data is drawn from a local distribution \( \rho_j \) on \( X \times Y \), with \( X \) as the input space and \( Y \) as the output space. The union of all local datasets forms the global dataset \( D = \bigcup_{j=1}^{m} D_j \), which follows a global distribution \( \rho \) on \( X \times Y \). The key objective of FL is to train a global model without sharing local data for privacy and communication efficiency. Ideally, the model would be trained on the entire dataset \( D \), but in practice, local data remains decentralized.

The number of examples on each local machine is denoted by \( n_j = |D_j| \), and the total number of examples is \( N = \sum_{j=1}^{m} n_j \). The goal is to train a global model using this decentralized data based on the empirical risk minimization objective. 
\subsection{Centralized Newton’s Method}

In centralized settings, the objective is commonly framed as:
\vspace{-3mm}
\begin{equation}
w_{D, \lambda} = \arg \min_{w \in \mathcal{H}_K} \left( \frac{1}{N} \sum_{i=1}^{N} \ell(f(x_i), y_i) + \lambda \alpha(w) \right),
\end{equation}

where \( \ell \) is the loss function, \( \alpha(w) \) is the regularization term, and \( \lambda > 0 \) is the regularization hyparameter. Here, we assume, \( f \in \mathcal{H}_K \) belongs to the reproducing kernel Hilbert space (RKHS).
 defined by a Mercer kernel \( K: X \times X \rightarrow \mathbb{R} \). This ensures that the model is defined as:
\begin{equation}
f(x) = \langle w, \phi(x) \rangle,
\label{eq:model_definition}
\end{equation}
where \( \phi : \mathcal{X} \rightarrow \mathcal{H}_K \) is the implicit feature mapping.

For convex and twice-differentiable loss functions, the optimization problem can be solved using the exact Newton method:

\vspace{-4.5mm}
\begin{equation}
w_{t+1} = w_t - \mu H_{D,t}^{-1} g_{D,t},
\label{eq:update_rule}
\end{equation}

Where \( \mu \) is the step size, and the gradient and Hessian at step \( t \) are:
\vspace{-3mm}
\[
g_{D,t} := \nabla L(D, w_t) + \lambda \nabla \alpha(w_t),
\]
\[
H_{D,t} := \nabla^2 L(D, w_t) + \lambda \nabla^2 \alpha(w_t).
\]

In federated learning, the system cannot access local training data directly, making it difficult to compute the global gradient \( g_{D,t} \) and Hessian matrix \( H_{D,t} \) accurately. The total time complexity of Newton's method is \( O(NM^2t) \), where \( M \) represents the dimension of the feature space and \( t \) denotes the number of iterations. Newton's method achieves a super-linear convergence rate of \( t = \log(\log(1/\delta)) \) for a \( \delta \)-accurate solution \cite{dennis1996numerical}.

\subsection{Newton’s Method with Partial Sketching}

To improve the computational efficiency, Newton Sketch \cite{pilanci2017newton} was introduced. This method constructs a structured random embedding of the Hessian matrix using a sketch matrix \( S \in \mathbb{R}^{k \times M} \), where \( k \ll N \). Partial Newton Sketch \cite{lacotte2021adaptive} improves this by sketching only the loss term while keeping the regularization term exact. The update rule becomes:
\vspace{-0.5mm}
\begin{align}
w_{t+1} &= w_t - \mu \tilde{H}_{D,t}^{-1} g_{D,t}, \quad \text{with} \nonumber \\
\tilde{H}_{D,t} &= \left( S \nabla^2 L(\mathcal{D}, w_t)^{1/2} \right)^\top \left( S \nabla^2 L(\mathcal{D}, w_t)^{1/2} \right) \nonumber \\
&\quad + \lambda \nabla^2 \alpha(w_t).
\end{align}

In this method, \( S \) is a zero-mean sketch matrix normalized such that \( \mathbb{E}[S^\top S / k] = I_M \). Different types of sketches, such as sub-Gaussian embeddings, SRHT \cite{ailon2006approximate}, and SJLT, offer trade-offs between sketching speed and communication complexity. SRHT achieves \( O(NM \log k) \) complexity, balancing fast computation and a small sketch size.

In federated settings, various Newton-type federated algorithms have been proposed to reduce communication overhead. Distributed Newton Local Newton \cite{gupta2021localnewton} applies Newton's method locally on each machine to speed up local convergence. FedNew \cite{elgabli2022fednew} uses a one-pass ADMM and an approximate Newton method to update the global model, while FedNL \cite{safaryan2022fednl} compresses local Hessian updates for efficient communication. SHED \cite{dal2024shed} leverages the eigendecomposition of local Hessians to reconstruct the global Hessian matrix, further reducing the communication burden.

\subsection{Federated Newton’s Method (FedNewton)}
To achieve an accurate approximation of the global model \( f_{D, \lambda} \), federated learning algorithms often rely on exchanging local information between clients. In methods like FedAvg and FedProx, this information typically consists of first-order gradients. However, we propose utilizing both first-order and second-order information to better capture the solution to equation (1) for the entire dataset \( D \).

It is important to observe that both the local gradients and Hessians in equation (3) can be aggregated into global gradient and Hessian representations. This motivates the development of the Federated Newton’s Method (FedNewton), which updates the model as follows:

\vspace{-3mm}

\begin{align}
w_{t+1} &= w_t - \mu H_{D,t}^{-1} g_{D,t}, \quad \text{with} \nonumber \\
H_{D,t} &= \sum_{j=1}^{m} \frac{n_j}{N} H_{D_j,t}, \quad g_{D,t} = \sum_{j=1}^{m} \frac{n_j}{N} g_{D_j,t}.
\end{align}

\subsubsection*{Complexity Analysis}
Before any iteration begins, computing the feature mapping on a local machine requires \( O(n_j M d) \) time. For the \( j \)-th local worker, the time complexity per iteration is at least \( O(n_j M^2 + n_j M) \) to compute the local Hessian \( H_{D_j,t} \) and gradient \( g_{D_j,t} \), respectively. Furthermore, summarizing the local Hessians and computing the global inverse Hessian incurs a time complexity of \( O(m M^2 + M^3) \).

However, the communication cost per iteration is \( O(M^2) \) due to the need to upload local Hessian matrices, making this approach impractical in many federated learning scenarios. Overall, the total computational complexity is \( O(\max_{j \in [m]} n_j M d + n_j M^2 t + M^3 t) \), and the communication complexity is \( O(M^2 t) \). Newton’s method achieves quadratic convergence with \( t = O(\log(\log(1/\delta))) \) for a \( \delta \)-approximation guarantee \cite{dennis1996numerical}, meaning \( L(D, w_t) - L(D, w_{D,\lambda}) \leq \delta \), where \( w_{D,\lambda} \) is the empirical risk minimizer as described in equation (1).
\vspace{-5.8mm}
\section{Methodology}
In this paper, we present \textbf{FLeNS (Federated Learning with Enhanced Nesterov-Newton Sketch)}, a novel federated learning algorithm that integrates Nesterov's accelerated gradient method \cite{nesterov1983method} with Hessian sketching \cite{pilanci2017newton} to enhance federated learning efficiency. Each client computes its local gradient and approximates its local Hessian using a sketching technique Subspace Randomized Hadamard Transform (SRHT) \cite{lacotte2021adaptive}. Instead of transmitting full Hessians, clients send these reduced-dimension sketched Hessians and gradients to a central server. The server aggregates the sketched Hessians and gradients to form global approximations. Leveraging Nesterov's momentum, the server updates the global model parameters using the aggregated data, performing accelerated gradient steps. This methodology reduces communication costs by minimizing transmitted data sizes and accelerates convergence by effectively utilizing second-order information in a compressed form.

\subsection*{Overview of Federated Optimization}

Federated learning involves optimizing a global model by distributing the computation across multiple clients, each of which holds a local dataset. The global objective function is to minimize the loss across all clients:
\vspace{-0.5mm}
\begin{equation}
\begin{aligned}
w^* &= \arg\min_{w \in \mathbb{R}^d} \left\{ L(w) = \frac{1}{N} \sum_{j=1}^m \sum_{i=1}^{n_j} \ell(f(w; x_{ij}), y_{ij}) \right. \\
&\quad \left. + \frac{\lambda}{2} \|w\|^2 \right\}
\end{aligned}
\end{equation}

where \(N = \sum_{j=1}^m n_j\) is the total number of data points across all clients, and \(\ell(f(w; x_{ij}), y_{ij})\) represents the loss function for data point \(x_{ij}, y_{ij}\).

\subsection*{Nesterov’s Accelerated Gradient}

The first key innovation in FLeNS is the integration of Nesterov’s acceleration at the client level. Nesterov's method provides an anticipatory momentum update, where the parameter \(v_t\) is computed based on both the current and the previous model states:
\begin{equation}
v_t = w_t + \beta_t (w_t - w_{t-1})
\label{eq:momentum_update}
\end{equation}
Here, \(\beta_t\) is a momentum term that controls the degree of acceleration. This update mechanism reduces the number of iterations required to reach a desired level of accuracy compared to standard gradient descent or Newton's method without acceleration, resulting in faster convergence rates. FLeNS leverages this accelerated approach to efficiently navigate the optimization landscape.

\subsection*{Hessian Sketching for Communication Efficiency}

A major bottleneck in federated learning is the communication overhead between clients and the central server. Second-order methods, while theoretically promising due to their use of curvature information (i.e., the Hessian), typically suffer from high communication costs, scaling as \(\mathcal{O}(M^2)\), where \(M\) is the dimensionality of the model. FLeNS addresses this limitation through Hessian sketching, which reduces the communication complexity by projecting the high-dimensional Hessian into a lower-dimensional space using a sketch matrix \(S_j \in \mathbb{R}^{k \times M}\), where \(k \ll M\):
\begin{equation}
\tilde{H}_{D_j,t} = S_j^\top \nabla^2 L_{D_j,t} S_j
\label{eq:sketched_hessian}
\end{equation}

This reduces the transmission complexity to \(\mathcal{O}(k^2)\), significantly improving the scalability of the algorithm. By applying Hessian sketching at each client, FLeNS reduces the amount of data exchanged during communication rounds, making the approach more practical for bandwidth-constrained networks.

\subsection*{Second-Order Information and Momentum Synergy}

While first-order methods (e.g., standard gradient descent) rely solely on gradient information for parameter updates, second-order methods utilize the Hessian to capture curvature information, leading to more informed updates. FLeNS maintains the advantages of second-order methods while simultaneously benefiting from the speed of Nesterov's acceleration. By combining Hessian sketching with Nesterov’s momentum, FLeNS achieves a synergy that enhances the optimization process:
\begin{equation}
w_{t+1} = w_t - \mu \tilde{H}_{D,t}^{-1} g_{D,t}
\label{eq:fednewton_update}
\end{equation}
where \(\tilde{H}_{D,t}\) is the aggregated sketched Hessian and \(g_{D,t}\) is the aggregated gradient. This update effectively balances the speed of convergence and the quality of updates, allowing the algorithm to navigate complex, non-convex loss landscapes more efficiently than algorithms relying solely on first-order gradients.

\subsection*{Communication and Aggregation}

At each iteration, clients compute their local gradients and sketched Hessians and send these to the central server. The server then aggregates these contributions:
\[
\tilde{H}_{D,t} = \sum_{j=1}^m \frac{n_j}{N} \tilde{H}_{D_j,t+1}, \quad g_{D,t} = \sum_{j=1}^m \frac{n_j}{N} g_{D_j,t+1}
\]
This aggregation ensures that the global model update reflects the collective information from all clients, making the optimization process robust to data heterogeneity across different clients.

\begin{algorithm}[h]
\small
\caption{FLeNS: Federated Learning with Enhanced Nesterov-Newton Sketch}
\textbf{Objective:} Optimize a global model using Nesterov’s acceleration combined with the sketching of the Hessian matrix in a federated learning setting.

\textbf{Setup:} Let $D_j$ represent the local dataset at client $j$ ($j = 1, \dots, m$). The global objective function is:
\[
w^* = \arg\min_{w \in \mathbb{R}^d} \left\{ L(w) = \frac{1}{N} \sum_{j=1}^m \sum_{i=1}^{n_j} \ell(f(w; x_{ij}), y_{ij}) + \frac{\lambda}{2} \|w\|^2 \right\}
\]
Where $N = \sum_{j=1}^m n_j$ is the total number of data points across all clients.

\textbf{Step 1: Local Gradient Computation and Sketching}\\
For each client $j \in \{1, \dots, m\}$:
\begin{itemize}
    \item Compute the local gradient $g_{D_j,t}$:
    \[
    g_{D_j,t} = \frac{1}{n_j} \sum_{i=1}^{n_j} \nabla_w \ell(f(w_t; x_{ij}), y_{ij}) + \lambda w_t
    \]
    \item Apply sketching to Hessian with sketch matrix $S_j \in \mathbb{R}^{k \times d}$:
    \[
    \tilde{H}_{D_j,t} = S_j^\top \nabla^2 L_{D_j,t} S_j
    \]
\end{itemize}

\textbf{Step 2: Nesterov’s Acceleration}\\
For each client $j$:
\begin{itemize}
    \item Apply Nesterov’s acceleration:
    \[
    v_t = w_t + \beta_t (w_t - w_{t-1})
    \]
    \item Update the gradient and sketched Hessian based on $v_t$:
    \[
    g_{D_j,t+1} = \nabla L(v_t), \quad \tilde{H}_{D_j,t+1} = \nabla^2 L(v_t)
    \]
\end{itemize}

\textbf{Step 3: Communication to Global Server}\\
For each client $j$:
\begin{itemize}
    \item Send $\tilde{H}_{D_j,t+1}$ and $g_{D_j,t+1}$ to the global server.
\end{itemize}

\textbf{Step 4: Aggregation at Global Server}\\
Aggregate the sketched Hessians and gradients:
\[
\tilde{H}_{D,t} = \sum_{j=1}^m \frac{n_j}{N} \tilde{H}_{D_j,t+1}, \quad g_{D,t} = \sum_{j=1}^m \frac{n_j}{N} g_{D_j,t+1}
\]

\textbf{Step 5: Global Model Update}\\
Update the global model parameters:
\[
w_{t+1} = w_t - \mu \tilde{H}_{D,t}^{-1} g_{D,t}
\]
Where $\mu$ is the step size.

\textbf{Step 6: Iterate}\\
Repeat steps 1-5 until convergence criteria are met.

\end{algorithm}


\subsection*{Computational Complexity}
The local gradient computation for \(n_j\) data points is \(O(n_j M)\), where \(n_j\) is the number of data points and \(M\) is the model dimension. Hessian sketching reduces this complexity to \(O(k n_j M + k^2 M)\), where \(k \ll M\) is the sketch size. At the server side, the per-iteration complexity involves aggregating the sketched Hessians from \(m\) clients at \(O(m k^2)\), gradients at \(O(m k)\), inverting the sketched Hessian at \(O(k^3)\), and updating the global model at \(O(M)\). Thus, the total computational complexity per iteration is \(O(n_j M)\) for each client and \(O(k^3 + m k^2 + M)\) at the server.

The communication complexity is minimized by transmitting sketched Hessians and gradients, with the communication cost from each client to the server being \(O(k^2)\) for the Hessians and \(O(k)\) for the gradients. The server sends back the updated global model at a cost of \(O(M)\), leading to a total communication complexity of \(O(k^2)\) from the client to the server and \(O(M)\) from the server to each client.

The sketch size \(k\) is critical for balancing computational efficiency and communication overhead. Choosing \(k \ll M\) reduces the size of the Hessians, thereby lowering both computational load and communication cost, while maintaining an effective approximation of the Hessian. FLeNS achieves a super-linear convergence rate of \(t = O(\log \log (1/\delta))\), where \(\delta\) represents the approximation accuracy.
\begin{table*}[ht]
\centering
\caption{Comparison of communication properties of FLeNS with previous algorithms}
\label{tab1}
\resizebox{\textwidth}{!}{
\begin{tabular}{|l|l|l|l|l|l|}
\hline
\textbf{Algorithm} & \textbf{Heterogeneous Setting} & \textbf{Sketch Size $k$} & \textbf{Iterations $t$} & \textbf{Communication per Round} & \textbf{Total Communication Complexity} \\ 
\hline
FedAvg \cite{li2020federated, su2021non} & Yes & - & $O\left( \frac{1}{\delta} \right)$ & $O(M)$ & $O\left( \frac{M}{\delta} \right)$ \\ 
\hline
FedProx \cite{lin2020optimal, su2021non} & Yes & - & $O\left( \frac{1}{\delta} \right)$ & $O(M)$ & $O\left( \frac{M}{\delta} \right)$ \\ 
\hline
DistributedNewton \cite{ghosh2020distributed} & No & - & $O\left( \log\left(\frac{1}{\delta}\right) \right)$ & $O(M)$ & $O\left( M \log\left(\frac{1}{\delta}\right) \right)$ \\ 
\hline
LocalNewton \cite{gupta2021localnewton} & No & - & $O\left( \log\left(\frac{1}{\delta}\right) \right)$ & $O(M)$ & $O\left( M \log\left(\frac{1}{\delta}\right) \right)$ \\ 
\hline
FedNL \cite{safaryan2022fednl} & Yes & - & $O\left( \log\left(\frac{1}{\delta}\right) \right)$ & $O(M)$ & $O\left( M \log\left(\frac{1}{\delta}\right) \right)$ \\ 
\hline
SHED \cite{fabbro2022newton} & Yes & - & $O\left( \log\left(\frac{1}{\delta}\right) \right)$ & - & $O(M^2)$ \\ 
\hline
FedNewton & Yes & - & $O\left( \log\log\left(\frac{1}{\delta}\right) \right)$ & $O(M^2)$ & $O\left( M^2 \log\log\left(\frac{1}{\delta}\right) \right)$ \\ 
\hline
FedNS \cite{li2024fedns} & Yes & $M$ & $O\left( \log\log\left(\frac{1}{\delta}\right) \right)$ & $O(kM)$ & $O\left( kM \log\log\left(\frac{1}{\delta}\right) \right)$ \\ 
\hline
FedNDES \cite{li2024fedns} & Yes & $\tilde{d} / \lambda$ & $O\left( \log\log\left(\frac{1}{\delta}\right) \right)$ & $O(kM)$ & $O\left( kM \log\log\left(\frac{1}{\delta}\right) \right)$ \\ 
\hline
FLeNS (Our Algorithm) & Yes & $O\left( \frac{N^\gamma}{2r + \gamma} \right)$ & $O\left( \log\log\left(\frac{1}{\delta}\right) \right)$ & $O(k^2)$ & $O\left( k^2 \log\log\left(\frac{1}{\delta}\right) \right)$ \\ 
\hline
\end{tabular}
}
\end{table*}


\section{Theoretical Guarantees}
In this section, we provide a detailed computational complexity analysis of the Federated Learning with Enhanced Nesterov-Newton Sketch (FLeNS) algorithm. The analysis is divided into two parts: the computational complexity at the client side and the computational complexity at the server side. Additionally, we examine the communication complexity between clients and the server.
\subsubsection*{Local Computations at Each Client}

Each client \( j \) performs computations to update local gradients and Hessians, apply Nesterov's acceleration, and prepare data for communication to the server.

\paragraph{Step 1: Local Gradient Computation} Each client computes the local gradient \( g_{D_j,t} \) at iteration \( t \):
\[
g_{D_j,t} = \frac{1}{n_j} \sum_{i=1}^{n_j} \nabla_w \ell \left( f \left( w_t; x_{ij} \right), y_{ij} \right) + \lambda w_t
\]
Cost per data point: \( O(M) \) operations per data point, where \( M \) is the dimension of model \( w \). Total cost per client: \( O(n_j M) \).

\paragraph{Step 2: Hessian Sketching} Each client computes a sketched Hessian using sketch matrix \( S_j \in \mathbb{R}^{k \times M} \), where \( k \ll M \):
\[
\tilde{H}_{D_j,t} = S_j^\top \nabla^2 L_{D_j,t} S_j
\]
Cost per Hessian-vector product: \( O(n_j M) \) per product. Total cost for \( k \) Hessian-vector products: \( O(k n_j M) \). Multiplying with \( S_j^\top \) requires \( O(k^2 M) \). Total cost for Hessian sketching:
\[
O(k n_j M + k^2 M)
\]

\paragraph{Step 3: Nesterov's Acceleration} Each client updates the momentum term \( v_t = w_t + \beta_t \left( w_t - w_{t-1} \right) \), costing \( O(M) \). Total cost per client: \( O(n_j M) \).

\subsubsection{Global Computations at the Server} The server aggregates sketched Hessians and gradients from all \( m \) clients:
\[
\tilde{H}_{D,t} = \sum_{j=1}^{m} \frac{n_j}{N} \tilde{H}_{D_j,t+1}, \quad g_{D,t} = \sum_{j=1}^{m} \frac{n_j}{N} g_{D_j,t+1}
\]
Cost: Aggregating gradients \( O(mk) \), Hessians \( O(m k^2) \). Inverting \( \tilde{H}_{D,t} \): \( O(k^3) \). Global model update: \( O(M) \). Total cost at the server:
\[
O(k^3 + m k^2 + M)
\]

\subsection{Communication Complexity}

\subsubsection{From Clients to Server} Clients send sketched Hessian \( \tilde{H}_{D_j,t+1} \in \mathbb{R}^{k \times k} \) and gradient \( g_{D_j,t+1} \in \mathbb{R}^k \), costing \( O(k^2) \) and \( O(k) \), respectively. Total communication: \( O(k^2) \).

\subsubsection{From Server to Clients} The server sends the updated model \( w_{t+1} \in \mathbb{R}^M \), costing \( O(M) \).

\subsection{Overall Complexity Summary}

\paragraph{Per Iteration Complexity:} Computational: Per client: \( O(n_j M) \), Server: \( O(k^3 + m k^2 + M) \). Communication: From each client: \( O(k^2) \), From server: \( O(M) \).

\paragraph{Total Complexity over \( T \) Iterations:} Computational: Per client: \( O(T n_j M) \), Server: \( O(T(k^3 + m k^2 + M)) \). Communication: From each client: \( O(T k^2) \), To each client: \( O(T M) \).

\subsection{Convergence Analysis of FLeNS}

In this section, we present a detailed convergence analysis of the Federated Learning with Enhanced Nesterov-Newton Sketch (FLeNS) algorithm. We outline the necessary assumptions and derive the convergence properties of the algorithm. The goal is to demonstrate that FLeNS converges to the global optimum under certain conditions, leveraging both Nesterov's acceleration and Hessian sketching.

\subsection*{Assumptions}

To establish the convergence of FLeNS, we make the following standard assumptions:

\paragraph{Assumption A1 (Smoothness of Loss Function):}
The loss function \( \ell(f(w;x),y) \) is twice continuously differentiable with respect to \( w \), and its gradient is Lipschitz continuous with constant \( L_1 > 0 \):
\[
\begin{aligned}
\|\nabla_w \ell(f(w;x),y) - \nabla_w \ell(f(v;x),y)\| &\leq L_1 \|w - v\|, \\
\forall w, v &\in \mathbb{R}^d, \forall x, y
\end{aligned}
\]

\paragraph{Assumption A2 (Strong Convexity and Smoothness of Loss Function \cite{pilanci2017newton}):}
The loss function \( \ell(f(w;x), y) \) is convex and twice continuously differentiable concerning \( w \). The gradient \( \nabla_w \ell(f(w;x),y) \) is Lipschitz continuous with constant \( L_1 > 0 \):
\[
\begin{aligned}
\|\nabla_w \ell(f(w;x),y) - \nabla_w \ell(f(v;x),y)\| &\leq L_1 \|w - v\|, \\
\forall w, v &\in \mathbb{R}^d
\end{aligned}
\]
The Hessian \( \nabla^2_w \ell(f(w;x),y) \) is Lipschitz continuous with constant \( L_2 > 0 \):
\[
\begin{aligned}
\|\nabla^2_w \ell(f(w;x),y) - \nabla^2_w \ell(f(v;x),y)\| &\leq L_2 \|w - v\|, \\
\forall w, v &\in \mathbb{R}^d
\end{aligned}
\]
\paragraph{Assumption A3 (Lipschitz Continuous Hessian \cite{boyd2004convex}):}
The Hessian \( \nabla^2 L(w) \) is Lipschitz continuous with constant \( L_2 > 0 \):
\[
\|\nabla^2 L(w) - \nabla^2 L(v)\| \leq L_2 \|w - v\|, \quad \forall w, v \in \mathbb{R}^d
\]

\paragraph{Assumption A4 (Bounded Hessian\cite{nesterov2013introductory}):}
The Hessian of the loss function is bounded:
\[
\|\nabla^2 \ell(f(w;x),y)\| \leq M, \quad \forall w \in \mathbb{R}^d, \forall x, y
\]

\paragraph{Assumption A5 (Bounded Variance of Stochastic Gradients \cite{nemirovski2009robust, bottou2018optimization}):}
The variance of the stochastic gradients is bounded:
\[
\mathbb{E}_{(x, y) \sim D} [\|\nabla_w \ell(f(w;x),y) - \nabla L(w)\|^2] \leq \sigma^2, \quad \forall w \in \mathbb{R}^d
\]

\paragraph{Assumption A6 (Quality of Hessian Sketching \cite{pilanci2017newton}):}
The sketch matrix \( S_j \in \mathbb{R}^{k \times d} \) satisfies the Subspace Embedding Property for the Hessian:
\[
\begin{aligned}
(1 - \epsilon) \|\nabla^2 L_{D_j,t} v\|^2 &\leq \|S_j^\top \nabla^2 L_{D_j,t} S_j v\|^2 \\
&\leq (1 + \epsilon) \|\nabla^2 L_{D_j,t} v\|^2, \quad \forall v \in \mathbb{R}^k
\end{aligned}
\]

where \( \epsilon \in (0, 1) \) is a small constant.

\paragraph{Assumption A7 (Parameters for Nesterov's Acceleration \cite{nesterov1983method}):}
The momentum parameter \( \beta_t \) is set according to:
\[
\beta_t = \frac{L_1 - \gamma}{L_1 + \gamma}
\]

\subsection*{Convergence Theorem}

\paragraph{Theorem 1 (Convergence of FLeNS):}
Under Assumptions A1–A7, if the step size \( \mu \) is chosen such that \( 0 < \mu \leq \frac{1}{L_1} \), the sequence \( \{w_t\} \) generated by FLeNS converges to the global optimum \( w^* \). Specifically, the convergence rate is given by:
\vspace{-3mm}
\[
L(w_t) - L(w^*) \leq \left( \frac{1}{1 + \kappa} \right)^{2t} (L(w_0) - L(w^*))
\]
where \( \kappa = \frac{L_1}{\gamma} \) is the condition number of the objective function.

\textit{Interpretation:}
\begin{itemize}
    \item The algorithm achieves a convergence rate of \( O\left(\left( \frac{1}{1 + \kappa} \right)^{2t} \right) \), which is faster than the standard gradient descent rate \( O\left(\left( 1 - \frac{1}{\kappa} \right)^t \right) \) and matches the accelerated gradient methods.
    \item The incorporation of Nesterov's acceleration and second-order information via Hessian sketching leads to an improved convergence rate.
\end{itemize}

\subsection*{Proof of Theorem 1}

The proof follows the standard analysis of Nesterov’s method combined with the sketching procedure. The steps are outlined below:

\paragraph{Step 1: Preliminaries}
Let \( e_t = w_t - w^* \) denote the error vector at iteration \( t \). From the strong convexity of \( L(w) \) (Assumption A2), we get:
\[
L(w_t) - L(w^*) \geq \frac{\gamma}{2} \|e_t\|^2
\]

\paragraph{Step 2: Update Equations}
The FLeNS algorithm updates are:
\[
v_t = w_t + \beta_t (w_t - w_{t-1}), \quad
w_{t+1} = w_t - \mu \tilde{H}_{D,t+1}^{-1} g_{D,t+1}
\]

\paragraph{Step 3: Error Recursion}
The error term \( e_{t+1} \) is expressed as:
\[
e_{t+1} = (1 - \mu) e_t - \mu \beta_t (e_t - e_{t-1})
\]

\paragraph{Step 4: Using the Hessian Approximation}
With the Hessian approximation (Assumption A6), the inverse approximation of the Hessian gives bounds on the error term, leading to the recursive bound:
\[
\|e_{t+1}\| \leq (1 - \mu) \|e_t\| + \mu \beta_t (\|e_t\| + \|e_{t-1}\|)
\]

\paragraph{Step 5: Solving the Recurrence}
This recurrence relation leads to the convergence rate, which is faster than the standard gradient descent and comparable to accelerated gradient methods:
\[
L(w_t) - L(w^*) \leq \left( \frac{1}{1 + \kappa} \right)^{2t} (L(w_0) - L(w^*))
\]

\subsection{Generalization Analysis of FLeNS}
To facilitate the Generalization analysis, we make use of the standard assumptions from A1 to A7:
\subsection*{Excess Risk Decomposition}

The excess risk is defined as the difference between the expected loss of the model obtained by FLeNS and that of the optimal model:
\[
E(w_t) = L(w_t) - L(w^*)
\]
We decompose the excess risk into two components:

\paragraph{Optimization Error (Empirical Risk Difference):}
\[
L_S(w_t) - L_S(w_S^*)
\]
where \( w_S^* = \arg \min_w L_S(w) \) is the minimizer of the empirical loss.

\paragraph{Generalization Error (Difference between Expected and Empirical Losses):}
\[
[L(w_t) - L_S(w_t)] + [L_S(w_S^*) - L(w^*)]
\]
Combining these, we have:
\begin{align}
E(w_t) &= [L_S(w_t) - L_S(w_S^*)] + [L(w_t) - L_S(w_t)] \notag \\
       &\quad + [L_S(w_S^*) - L(w^*)]
\end{align}

Under Assumption G1, \( L_S(w_S^*) \leq L_S(w^*) \), so \( L_S(w_S^*) - L(w^*) \leq L_S(w^*) - L(w^*) \).

\subsection*{Bounding the Optimization Error}

\paragraph{Theorem 1 (Optimization Error Bound):}
Under Assumptions G2 and G3, the optimization error after \( t \) iterations of FLeNS is bounded by:
\[
L_S(w_t) - L_S(w_S^*) \leq \left( \frac{1}{1 + \kappa} \right)^{2t} \left( L_S(w_0) - L_S(w_S^*) \right)
\]
where \( \kappa = \frac{L_1}{\gamma} \) is the condition number of the empirical loss function \( L_S(w) \).

\textit{Proof Sketch:} The result follows from the convergence analysis of FLeNS, where we have established that the sequence \( \{w_t\} \) converges to \( w_S^* \) at an accelerated rate due to Nesterov's acceleration. The convergence rate is characterized by the factor \( \left( \frac{1}{1 + \kappa} \right)^{2} \), leading to the exponential decay of the optimization error.

\subsection*{Bounding the Generalization Error}

\paragraph{Lemma 1 (Generalization Error of Empirical Minimizer):}
Under Assumptions G1–G4, with probability at least \( 1 - \delta \), the following holds:
\[
L(w_S^*) - L(w^*) \leq \frac{2G^2}{\gamma N} + \frac{G^2 \log(1/\delta)}{N}
\]

\paragraph{Lemma 2 (Generalization Error of Model at Iteration \( t \)):}
Under Assumptions G1–G4, with probability at least \( 1 - \delta \), we have:
\[
L(w_t) - L_S(w_t) \leq \frac{G^2 \log(1/\delta)}{N}
\]


\subsection*{Total Excess Risk Bound}

\paragraph{Theorem 2 (Total Excess Risk Bound):}
Under Assumptions G1–G5, with probability at least \( 1 - 2\delta \), the excess risk of the model \( w_t \) obtained by FLeNS satisfies:

\begin{align}
E(w_t) \leq \left( \frac{1}{1 + \kappa} \right)^{2t} 
\left( L_S(w_0) - L_S(w_S^*) \right) &+ \frac{2G^2}{\gamma N} \notag \\
&+ \frac{2G^2 \log(1/\delta)}{N}
\end{align}

\section{Compared with Related Work}
Table \ref{tab1} reports the computational properties of related work to achieve a $\delta$-accurate solution in terms of the regularized least squares loss. We compare our proposed algorithm \textbf{FLeNS} with first-order methods and other Newton-type federated learning (FL) algorithms.

FedNS applies commonly used sketching approaches, such as sub-Gaussian random projections, SRHT, and Sparse Johnson-Lindenstrauss Transform (SJLT). FLeNS primarily utilizes SRHT due to its favourable computational properties. Different sketch types lead to various sketch times on the $j$-th local machine: $O(n_j M k)$ for Sub-Gaussian,
   $O(n_j M \log k)$ for  SRHT,
 $O(\text{nnz}(\phi(X_j)))$ for SJLT.

\textbf{Compared with first-order algorithms }federated Newton's methods, including FedNS, FedNDES, and our proposed FLeNS, converge much faster ($O(\log \frac{1}{\delta})$ or $O(\log \log \frac{1}{\delta})$ vs. $O(\frac{1}{\delta})$ for first-order methods like FedAvg and FedProx). However, the communication complexities of federated Newton's methods are generally higher, at least $O(k M t)$, while it is $O(M t)$ for first-order methods (FedAvg and FedProx).


\textbf{Compared with Newton-type FL methods} Distributed Newton \cite{ghosh2020distributed} and Local Newton \cite{gupta2021localnewton} perform Newton's method using local information and implicitly assume that local datasets across devices are homogeneous, which limits their applicability in federated learning. In contrast, our proposed algorithm communicate local sketched Hessian matrices to approximate the global Hessian, making them naturally applicable to heterogeneous settings.

\subsection*{Recent Newton-Type FL Methods}
Several recent Newton-type FL methods include the following: FedNL \cite{safaryan2022fednl} compresses the difference between the local Hessian and the global Hessian from the previous step and transfers the compressed difference to the global server for merging. Theoretically, FedNL achieves at least linear convergence \( O(\log \frac{1}{\delta}) \) with a communication cost of \( O(M) \) per round. FedNew \cite{elgabli2022fednew} uses ADMM to solve an unconstrained convex optimization problem for obtaining the local update directions \( H_{D_j,t-1} g_{D_j,t} \) and performs Newton's method by averaging these directions at the server. However, this method only proves convergence without explicit convergence rates. SHED \cite{fabbro2022newton} performs eigendecomposition on the local Hessian and incrementally sends the eigenvalues and eigenvectors to the server, where the local Hessians are reconstructed to perform Newton's method, achieving super-linear convergence with a total communication cost of \( O(M^2) \). FedNS \cite{li2024fedns} utilizes Hessian sketching to approximate the global Newton step. By communicating sketched Hessians of size \( k \times M \) from each client to the server, FedNS reduces communication costs compared to transmitting full Hessians. It achieves super-linear convergence rates \( O(\log \log \frac{1}{\delta}) \), although the communication per round is \( O(k M) \), which can still be significant if \( k \) is large. FedNDES \cite{li2024fedns} improves upon FedNS by using dimension-efficient sketching and adaptive strategies to further reduce the sketch size \( k \) to \( \tilde{d}_{\lambda} \), the empirical effective dimension of the data, resulting in lower communication costs per round \( O(k M) \) while maintaining the super-linear convergence rate \( O(\log \log \frac{1}{\delta}) \).

\textbf{FLeNS (our proposed algorithm):} Further advances the state-of-the-art by integrating Nesterov's acceleration with Leveraging dimension-efficient Hessian sketching and a smaller sketch size $k$. FLeNS achieves accelerated convergence rates while significantly reducing the communication per round to $O(k^2)$, thanks to the smaller sketch size and the efficient combination of momentum and second-order information.

\vspace{-2.9mm}
 \section{Experiments}
In this section, we conduct experiments to validate our theoretical claims using several real-world federated datasets. Our implementations are based on publicly available code from FedNew \cite{elgabli2022fednew} and FedNL \cite{safaryan2022fednl}. However, we excluded the SHED algorithm due to the unavailability of public code \cite{fabbro2022newton}. The base model for all methods is logistic regression, and the algorithms update the Hessian matrix at each iteration.

We begin by analyzing the effect of sketch size on the proposed FLeNS algorithm, followed by comparing related methods in terms of communication rounds. As in FedNew \cite{elgabli2022fednew}, we consider regularized logistic regression with the loss function \( L(D, w) = \frac{1}{N} \sum_{i=1}^{N} \log(1 + \exp(-y_i x_i^\top w)) + \lambda \|w\|^2_2 \), where \( \lambda \) is the regularization parameter, set to \( 10^{-3} \) for all experiments. Each experiment was conducted 10 times, and the results were averaged. The figures \ref{fig1}, \ref{fig2}, and \ref{fig3}
 represent the mean performance. We use the optimality gap, \( L(w_t) - L(w^*) \), as the primary performance metric, with \( w^* \) being the solution estimated by global Newton’s method.

The algorithms were evaluated using publicly available LIBSVM datasets \cite{chang2011libsvm}. The dataset statistics and relevant hyperparameters are summarized in Table \ref{tab2}. In particular, the hyperparameters $\rho$ and $\alpha$ are used in the FedNew algorithm.
\begin{table}[h!]
    \centering
    \vspace{0.3cm} 
    \small 
    \begin{tabular}{|l |l |l |l |l |l |l|}
        \hline
        Dataset & $n$ & $M$ & $k$ & $m$ & $\rho$ & $\alpha$ \\
        \hline
        phishing & 11,055 & 68 & 17 & 40 & 0.1 & 0.25 \\
        covtype & 581,012 & 54 & 20 & 200 & 50 & 1 \\
        SUSY & 5,000,000 & 18 & 10 & 1000 & 50 & 1 \\
        \hline
    \end{tabular}
    \vspace{0.3cm} 
    \caption{Summary of datasets and hyperparameters.}
    \label{tab2}
\end{table}

\par

\section{Results and Discussions}

\subsection*{\textbf{Convergence comparison.}} Figure \ref{fig1} shows the loss discrepancy (the difference between the loss of the trained model and the optimal solution) over the number of communication rounds for various methods reporting the convergence of the compared methods, and highlighting several key findings. 1) There are significant gaps between the convergence speeds of the proposed method, FLeNS, and existing Newton-type federated learning (FL) methods such as FedNS, FedNDES, FedNew, FedNL, and FedAvg. These gaps validate the accelerated super-linear convergence of FLeNS over other methods. 2) FLeNS converges faster than both FedNS and FedNDES, approaching the performance of FedNewton while requiring less communication and computation. In contrast, FedNew and FedNL converge more slowly, closely resembling the behaviour of FedAvg. 3) Since a lower loss generally correlates with higher predictive accuracy on the test set, we can conclude that FLeNS results in better final predictive accuracy. Despite small sketch sizes, FLeNS demonstrates robustness by maintaining considerable accuracy, making it suitable for scenarios with limited computational resources. 4) FLeNS communicates sketched Hessians of size $k \times k$, denoted as $O(\tilde{d} \lambda^2)$, significantly reducing per-round communication compared to methods like FedNS and FedNDES. Additionally, Nesterov's acceleration allows FLeNS to converge faster, meaning it requires fewer iterations to achieve the desired accuracy. By combining lower per-round communication costs with fewer communication rounds, FLeNS results in the lowest total communication cost among the three methods.

\subsection*{\textbf{Impact of sketch size on Performance.}} Figure \ref{fig2} reports the results of predictive accuracy as a function of sketch size and highlights several important observations:
1) Increasing the sketch size consistently improves generalization performance. 2) FLeNS algorithms eventually converge towards the global Newton’s method. 3) Even with a small sketch size, i.e., \( k \ll M \) and \( k \ll N \), the performance remains strong.

\begin{figure}
    \centering
    \includegraphics[width=\linewidth]{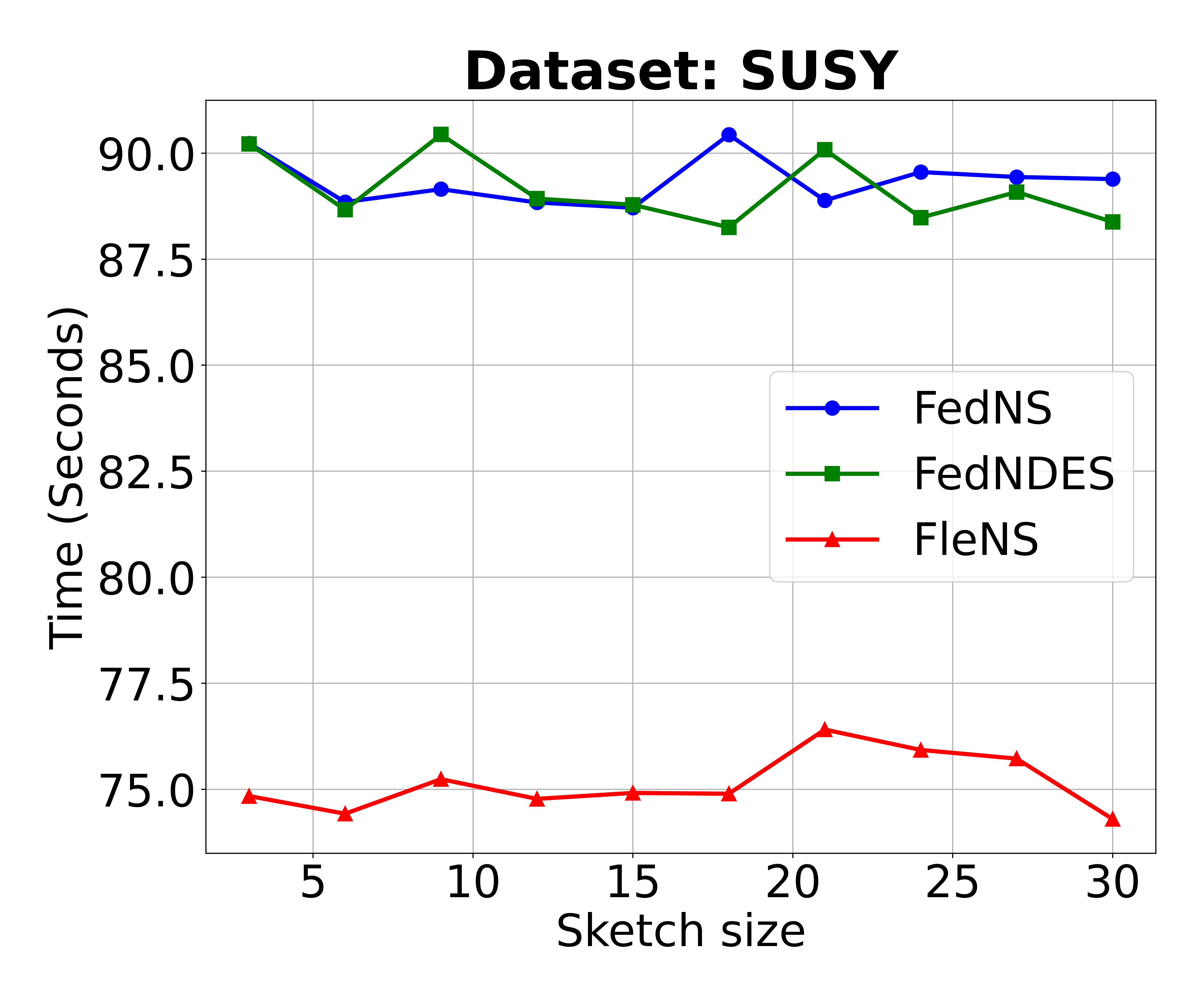} 
    \caption{Computational time wrt Sketch size.}
    \label{fig3}
\end{figure}

\subsection*{\textbf{Impact of sketch size on Computational Time.}} Figure \ref{fig3} shows that FLeNS is significantly more efficient in terms of computational time compared to other methods like FedNS and FedNDES, particularly as the sketch size increases.
The plot demonstrates the advantage of using FLeNS, especially when dealing with large sketch sizes, as it maintains low computation time while the other methods experience higher and more variable times.

\vspace{-1mm}
\section*{Conclusion}
In this paper, we introduced FLeNS (Federated Learning with Enhanced Nesterov-Newton Sketch), a highly efficient and scalable optimization framework that addresses the inherent limitations of both first-order methods and traditional Newton-type algorithms in federated learning. By integrating Nesterov's acceleration with dimension-efficient Hessian sketching, FLeNS harnesses the power of second-order information while significantly reducing communication overhead. This novel combination enables FLeNS to achieve super-linear convergence rates, surpassing existing federated Newton-type algorithms such as FedNS, FedNew, and FedNDES.

Unlike conventional methods that require transmitting full Hessian matrices, FLeNS reduces the dimensionality of the Hessian, striking an optimal balance between computational efficiency and communication cost. Its robustness in bandwidth-constrained and privacy-sensitive environments further demonstrates its adaptability, making it a versatile solution across a diverse range of federated learning applications. Furthermore, FLeNS excels at balancing key trade-offs between sketch size, communication efficiency, and convergence speed, reinforcing its practical utility in real-world scenarios where both communication constraints and data heterogeneity are prominent challenges.

\bibliographystyle{unsrt}  
\bibliography{FLeNS}

\end{document}